\documentclass[a4paper]{llncs}
\usepackage{cmap}
\usepackage{amsmath,amsfonts,amssymb}
\usepackage{mathtools}
\usepackage{graphicx}
\usepackage{microtype}
\usepackage{colonequals,xspace}
\usepackage[ruled,vlined]{algorithm2e}
\usepackage{wrapfig}
\usepackage{subfig}
\usepackage{booktabs}
\usepackage{nicefrac}
\usepackage{tikz}
\usepackage{multirow}
\usepackage{algorithm2e}
\usepackage{todonotes}
\usepackage[]{algorithm2e}
\usepackage{adjustbox}

\usetikzlibrary{arrows,decorations.pathmorphing,positioning,fit,trees,shapes,shadows,automata,calc}
\DontPrintSemicolon
\SetKwInOut{Input}{Input}\SetKwInOut{Output}{Output}

\SetCommentSty{mycommfont}

\tikzset{>=stealth}
\tikzset{state/.append style={inner sep=2pt,minimum size=2pt}}

\usepackage{array}
\newcolumntype{H}{>{\setbox0=\hbox\bgroup}c<{\egroup}@{}}

\newcommand{\ie}{i.\,e.,\@\xspace}

\newcommand{\eg}{e.\,g.,\@\xspace}

\newcommand{\prismpomdp}{\textrm{PRISM}-pomdp\xspace}

\newcommand{\tool}[1]{\textrm{#1}\xspace}
\DeclareMathOperator*{\argmax}{argmax}

\DeclareMathOperator{\dom}{dom}

\DeclareMathOperator{\successors}{succ}

\newcommand{\mdp}{\mathsf{M}}
\newcommand{\dtmc}{\mathsf{C}}

\newcommand{\pomdp}{\mathsf{D}}
\newcommand{\acts}{\mathit{Act}}
\newcommand{\act}{\alpha}
\newcommand{\sinit}{s_\text{init}}

\newcommand{\states}{S}
\newcommand{\dists}{\mathit{Dist}}
\newcommand{\dist}{\mu}
\newcommand{\prob}{P}

\newcommand{\rew}{R}

\newcommand{\sched}{\sigma}
\newcommand{\schedset}{\Sigma}

\newcommand{\obss}{\mathcal{O}}
\newcommand{\obs}{\lambda}

\newcommand{\pathset}{\mathsf{Paths}}

\newcommand{\pathsfin}{\pathset_{\mathit{fin}}}
\newcommand{\pathsinf}{\pathset_{\mathit{inf}}}

\newcommand{\last}[1]{\mathit{last}(#1)}

\newcommand{\dcup}{\mathbin{\dot\cup}}
\newcommand{\partialto}{\mathrel{\ooalign{\hfil$\mapstochar\mkern5mu$\hfil\cr$\to$\cr}}}

\DeclareMathAlphabet{\mathpzc}{OT1}{pzc}{m}{it}
\def\presuper#1#2%
  {\mathop{}%
   \mathopen{\vphantom{#2}}^{#1}%
   \kern-\scriptspace%
   #2}

\begin{document}

\author{Leonore Winterer\thanks{Corresponding author}\inst{1} \and Ralf Wimmer\inst{2,1}\and Nils Jansen\inst{3}
   \and Bernd Becker\inst{1}}
\title{Strengthening Deterministic Policies \\ for POMDPs}
\institute{
  Albert-Ludwigs-Universit\"at Freiburg, Freiburg im Breisgau, Germany \\
  \texttt{\{winterel, wimmer, becker\}@informatik.uni-freiburg.de}
  \and
  Concept Engineering GmbH, Freiburg im Breisgau, Germany\\
  \and
  Radboud University, Nijmegen, The Netherlands \\
  \texttt{n.jansen@science.ru.nl}
}
\authorrunning{L.\ Winterer \emph{et al.}}
\makeatletter
\titlerunning{\@title}
\makeatother
\maketitle

\begin{abstract}
The synthesis problem for partially observable Markov decision processes (POMDPs) is to compute a policy that satisfies a given specification.
Such policies have to take the full execution history of a POMDP into account, rendering the problem undecidable in general.
A common approach is to use a limited amount of memory and randomize over potential choices.
Yet, this problem is still NP-hard and often computationally intractable in practice.
A restricted problem is to use neither history nor randomization, yielding policies that are called stationary and deterministic.
Previous approaches to compute such policies employ mixed-integer linear programming (MILP).
We provide a novel MILP encoding that supports sophisticated specifications in the form of temporal logic constraints.
It is able to handle an arbitrary number of such specifications.
Yet, randomization and memory are often mandatory to achieve satisfactory policies.
First, we extend our encoding to deliver a restricted class of randomized policies.
Second, based on the results of the original MILP, we employ a preprocessing of the POMDP to encompass memory-based decisions.
The advantages of our approach over state-of-the-art POMDP solvers lie
(1)~in the flexibility to strengthen simple deterministic policies without losing computational tractability and
(2)~in the ability to enforce the provable satisfaction of arbitrarily many specifications.
The latter point allows to take trade-offs between performance and safety aspects of typical POMDP examples into account.
We show the effectiveness of our method on a broad range of benchmarks.
\end{abstract}

\section{Introduction}
\label{sec:introduction}
Partially observable Markov decision processes (POMDPs) are a formal model for planning under uncertainty in partially observable environments~\cite{kaelbling1998planning,thrun2005probabilistic}.
POMDPs adequately model a number of real-world applications, see for instance~\cite{WongpiromsarnF12,DBLP:books/daglib/0023820}.
While an agent operates in a scenario modeled by a POMDP, it receives \emph{observations} and tries to infer the likelihood of the system being in a certain state, the belief state.
Together with a belief update function, the space of all belief states forms a (uncountably infinite) \emph{belief MDP}~\cite{ShaniPK13,MadaniHC99,braziunas2003pomdp}.

\begin{figure}[t]
	\centering
	\scalebox{0.75}{  \begin{tikzpicture}[>=stealth]
    \node[draw, align=center] (s1) at (0,0) [draw,thick,minimum width=2cm,minimum height=1cm] {Docked in\\left station};
    \node[draw, align=center] (s2) at (3,1) [draw,thick,minimum width=2cm,minimum height=1cm] {Facing\\left station};
    \node[draw, align=center, fill=yellow!20] (s3) at (3,-1) [draw,thick,minimum width=2cm,minimum height=1cm] {Back to\\left station};
    \node[draw, align=center, fill=yellow!20] (s4) at (6,1) [draw,thick,minimum width=2cm,minimum height=1cm] {Space,\\facing left};
    \node[draw, align=center, fill=yellow!20] (s5) at (6,-1) [draw,thick,minimum width=2cm,minimum height=1cm] {Space,\\facing right};
    \node[draw, align=center] (s6) at (9,-1) [draw,thick,minimum width=2cm,minimum height=1cm] {Facing\\right station};
    \node[draw, align=center, fill=yellow!20] (s7) at (9,1) [draw,thick,minimum width=2cm,minimum height=1cm] {Back to\\right station};
    \node[draw, align=center] (s8) at (12,0) [draw,thick,minimum width=2cm,minimum height=1cm] {Docked in\\right station};

    \draw[->] (4,1.25) -- (5,1.25);
    \draw[->] (5,0.75) -- (4,0.75);
    
    \draw[->] (7,1.25) -- (8,1.25);
    \draw[->] (8,0.75) -- (7,0.75);
    
    \draw[->] (4,-1.25) -- (5,-1.25);
    \draw[->] (5,-0.75) -- (4,-0.75);
    
    \draw[->] (7,-1.25) -- (8,-1.25);
    \draw[->] (8,-0.75) -- (7,-0.75);
    
    \draw[->] (1, 0.25) -- (2, 1.25);
    \draw[->] (2, 1) -- (1, 0);
    
    \draw[->] (1, -0.25) -- (2, -1.25);
    \draw[->] (2, -1) -- (1, 0);
    
    \draw[->] (11, 0.25) -- (10, 1.25);
    \draw[->] (10, 1) -- (11, 0);
    
    \draw[->] (11, -0.25) -- (10, -1.25);
    \draw[->] (10, -1) -- (11, 0);

    
    
\end{tikzpicture}}
	\caption{The space shuttle benchmark -- yellow states share an observation. Transitions have been simplified for clarity.}
	\label{fig:shuttle}
\end{figure}
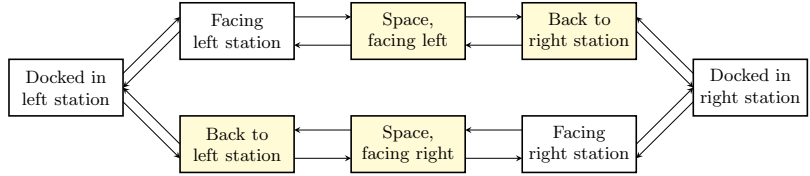

Consider the following simple example~\cite{DBLP:conf/aaai/Chrisman92} as sketched in Fig.~\ref{fig:shuttle}.
A space shuttle has to transport goods between two stations, while docking at these stations is subject to failure with certain probabilities. 
The perception of the shuttle is limited in the sense that it will only see the stations if it is directly facing them. 
If not, it can only see empty space and has to infer from the history which of the stations is the next one to deliver goods to.

Traditional POMDP problems typically comprise the computation of a policy that maximizes a cumulative reward over a finite horizon.
However, the application may require that the agent's behavior obeys more complicated specifications.
For example, temporal logics (\eg~LTL~\cite{Pnueli77}) describe task properties like reachability or liveness that cannot be expressed using reward functions~\cite{littman2017environment}.
For the aforementioned space shuttle, maximizing the reward corresponds to maximizing the number of succesful deliveries. 
Additional specifications may for instance require the shuttle to only navigate in empty space for a limited number of steps.

Policy synthesis for POMDPs is hard.
For infinite- or indefinite-horizon problems, computing an optimal policy is undecidable~\cite{MadaniHC99}.
Optimal action choices depend on the whole history of observations and actions, and thus require an infinite amount of memory.
When restricting the specifications to maximizing accumulated rewards over a finite horizon and also limiting the available memory, computing an optimal policy is PSPACE-complete~\cite{papadimitriou1987complexity}.
This problem is, practically, intractable even for small instances~\cite{meuleau1999learning}.
When policies are restricted to be \emph{memoryless}, finding an optimal policy within this set is still NP-hard~\cite{VlassisLB12}.
For the more general LTL specifications, synthesis of policies with limited memory is even EXPTIME-complete~\cite{chatterjee2015qualitative}.

\paragraph{State-of-the-art.}
The aforementioned hardness and intractability of the computation of exact solutions for the POMDP problems discussed earlier triggered several feasible approaches.
Notably, there are approximate~\cite{hauskrecht2000value}, point-based~\cite{pineau2003point}, or Monte-Carlo-based~\cite{silver2010monte} methods.
Yet, none of these approaches provides guarantees for temporal logic specifications.
The tool \prismpomdp~\cite{NPZ17} actually provides guarantees by approximating the belief space into a fully observable belief MDP, but is restricted to small examples.
Other techniques, such as those employing an incremental satisfiability modulo theory (SMT) solver over a bounded belief space~\cite{wang2018bounded} or a simulation over sets of belief models \cite{haesaert2018temporal}, are also restricted to small examples.
\cite{DBLP:conf/cdc/WintererJW0TK017} employs a game-based abstraction approach to efficiently solve problems with specific properties.
In~\cite{junges2018finite}, finite-state controllers for POMDPs are computed using parameter synthesis for Markov chains~\cite{param_sttt,DBLP:journals/corr/abs-1903-07993} by applying convex optimization techniques~\cite{DBLP:conf/tacas/Cubuktepe0JKPPT17,DBLP:conf/atva/CubuktepeJJKT18}.
Another work employs machine learning techniques together with formal verification to achieve sound but not optimal solutions~\cite{DBLP:conf/ijcai/Carr0WS0T19}.

\paragraph{Our Approach.}
The problem we consider in this paper is to compute a policy for a POMDP that provably satisfies one or more specifications such as temporal logic constraints and expected (discounted) reward properties.
First, we restrict ourselves to a simple class of policies which are both memoryless and do not randomize over action choices, that is, they are \emph{deterministic}. 
A natural approach encodes this problem as a mixed-integer linear program (MILP)~\cite{DBLP:books/daglib/0090562}.
We extend previous approaches~\cite{DBLP:conf/aips/ArasDC07,DBLP:conf/aips/KumarMZ16} to account for multiple specifications and provide a particular encoding for temporal logic constraints.
The advantage is that these MILPs yield simple, small, and easy-to-analyze policies which can be computed by efficient state-of-the-art tools like \tool{Gurobi}~\cite{gurobi}. 

However, policies that incorporate randomization over choices often trade off the necessity of memory-based decisions~\cite{chatterjee2004trading,amato2010optimizing}, and randomization may be needed for multiple objectives~\cite{DBLP:journals/lmcs/EtessamiKVY08,DBLP:conf/csl/BaierDK14}.
To preserve the advantages of MILP solving, we propose \emph{static randomization}. 
We augment the MILP encoding in the following way. 
In addition to deterministic choices, the policy may to select an arbitrary but fixed distribution over all possible actions.
As we will demonstrate in this paper, often any distribution is sufficient as long as randomization is possible.

Yet, for certain problems a notion of (at least finite) memory is required. 
As a third step to strengthen deterministic policies, we perform a preprocessing of the POMDP regarding previous computations for purely deterministic policies. 
At states where the choices are bad according to the specifications, we perform a technique we call \emph{observation and state splitting} which essentially encodes finite memory into the state space of the POMDP.
Intuitively, we enable a policy to distinguish states that previously shared an observation. 

Summarized, we provide three contributions. 
First, we enable the computation of deterministic polices that provably adhere to multiple specifications. 
Second, we augment the underlying MILP to account for randomization using fixed distributions over actions.
Third, we introduce a novel POMDP preprocessing which encodes finite memory into critical states. 
We showcase the feasibility and competitiveness of our approach by a thorough experimental evaluation on well-known case studies.

\section{Preliminaries}
\label{sec:preliminaries}

For a finite or countably infinite set $X$, $\dist\colon X\to[0,1]$ with $\sum_{x\in X}\dist(x) = 1$ denotes a \emph{probability distribution} over $X$;
the set of all probability distributions over $X$ is $\dists(X)$.
A partial function $f\colon X \partialto Y$ is a function $f\colon X' \rightarrow Y$ for some subset $X'=\dom(f) \subset X$.

\begin{definition}[Markov Decision Process]
  \label{def:mdp}
  A \emph{Markov Decision Process (MDP)} is a tuple $\mdp = (\states, \sinit, \acts, \prob, \rew)$ where
  $\states$ is a finite set of \emph{states},
  $\sinit\in\states$ the \emph{initial state},
  $\acts$ a finite set of \emph{actions},
  $\prob\colon \states\times\acts\partialto\dists(\states)$ a (partial) \emph{probabilistic transition function},
  and $\rew\colon \states\times\acts\to \mathbb{R}$ a reward function that assigns to every tuple $(s,\act)\in\dom(\prob)$ a
  real-valued reward.
\end{definition}
The set of actions that are enabled in $s$ is $\acts(s) = \{\act\in\acts\,|\,(s,\act)\in\dom(\prob)\}$ .

A partially observable Markov decision process (\emph{POMDP})~\cite{kaelbling1998planning} models restricted knowledge of the current state of an MDP.
\begin{definition}[POMDP]
  \label{def:pomdp}
  A \emph{partially observable Markov decision process (PO\-MDP)} is a tuple
  $\pomdp = (\mdp,\obss,\obs)$
  such that
  $\mdp=(\states,\sinit,\acts,\prob,\rew)$ is the \emph{underlying MDP of $\pomdp$}, $\obss$ a finite set of
  \emph{observations}, and $\obs\colon \states\to\obss$ the \emph{observation function}.
\end{definition}
Note that in our definition of a POMDP each state has exactly one observation.
Sometimes a more general definition of POMDPs is used, in which the observation function depends not only
on the current state, but also on the previous action, and returns not a fixed observation, but a probability distribution over the possible observations.
However, there is a polynomial reduction from this general case to the one we use in this work~\cite{ChatterjeeCGK16}.

\begin{definition}[Path]
 \label{def:path}
 A sequence $\pi = s_0\act_0s_1\act_1\dots$ with $s_i\in \states$, $\act_i \in \acts$ and $\prob(s_i,\act_i)(s_{i+1})>0$
 for all $i\geq 0$ is called a \emph{path}. 
 Paths can be finite (ending in a state) or infinite. 
 The set of finite paths is $\pathsfin$ and the set of infinite paths $\pathsinf$. For a finite path $\pi$, 
 we denote by $\last{\pi}$ the final state of $\pi$.
\end{definition}

\begin{definition}[Observation Sequence]
 \label{def:history}
 If $\pi = s_0\act_0s_1\act_1\dots s_{n-1}\act_{n-1}s_n$ is a finite path,
 then $\theta \coloneqq \lambda(\pi) \coloneqq \lambda(s_0)\act_0\lambda(s_1)\act_1\dots \lambda(s_{n-1})\act_{n-1}\lambda(s_n)$ 
 is called the \emph{observation sequence} of $\pi$.
\end{definition}
Before a probability space over paths of (PO)MDPs can be defined, the nondeterminism needs to be
resolved. 
This resolution is done by an entity called a policy that determines the next action to
execute:

\begin{definition}[Policy]
 \label{def:policy}
  A \emph{policy} for a POMDP $\pomdp$ is a function $\sigma\colon\pathsfin\to\dists(\acts)$ such that $\sigma(s_0\ldots s_n)(\act)>0$ implies $\act\in\acts(s_n)$.
  We denote the set of all possible policies for a POMDP $\pomdp$ with $\schedset_{\pomdp}$.

  A policy is \emph{observation-based} if
  $\sigma(\pi)=\sigma(\pi')$ holds for all $\pi,\pi'$ with $\lambda(\pi)=\lambda(\pi')$.
  A policy is $\sigma$ \emph{stationary} if $\sigma(\pi)=\sigma(\pi')$ holds whenever
  $\last{\pi}=\last{\pi'}$. 
  The set of all stationary policies for a POMDP $\pomdp$ is $\schedset^\text{stat}_{\pomdp}$.
  Policies that are not stationary, are called \emph{history-dependent}.
  A policy is \emph{deterministic} if $\sigma(\pi)(\act)\in\{0,1\}$
  for all $\pi$ and $\act$. Policies that are not deterministic are called \emph{randomized}.
\end{definition}
Stationary observation-based policies are typically regarded as functions $\sigma\colon\obss\to\dists(\acts)$ (randomized policy) or
$\sigma\colon\obss\to\acts$ (deterministic policy). 
As a policy resolves all nondeterminism and partial observability, it turns a (PO)MDP into a discrete-time Markov chain (DTMC), which is a purely stochastic process.
\begin{definition}[Induced DTMC]
  Let $\pomdp$ be a POMDP as defined above with reward function $\rew$
  and $\sigma\colon\pathsfin\to\dists(\acts)$ a policy.
  The induced DTMC is a tuple $\pomdp_\sigma = (\pathsfin, \sinit, P')$ such that
  $\prob'(\pi,\pi') = \sigma(\pi)(\act)\cdot \prob(\last{\pi},\act,s)$
  if $\pi' = \pi\act s$, and $P'(\pi,\pi') = 0$ otherwise.
  The induced reward function $\rew'\colon\pathsfin\to\mathbb{R}$ is defined
  as $\rew'(\pi) = \sum_{\act\in\acts(\last{\pi})} \sched(\pi)(\act)\cdot\rew(\last{\pi},\act)$.
\end{definition}
In the following we consider the computation of observation-based policies for POMDPs.
The goal is to find a policy such that the induced DTMC satisfies a given specification.
For the scope of this paper, we focus on \emph{reachability} and \textit{expected discounted reward}
specifications and combinations thereof.
Note that general LTL properties for probabilistic systems can be reduced to reachability~\cite{BK08}.

\begin{definition}[Reachability]
  \label{def:reachability}
  Let $\dtmc=(S,\sinit,\prob)$ be a DTMC and $T\subseteq S$ a set of target states. The probability
  to reach a state in $T$ from $s$ is the unique solution of the following linear equation system:
  \[
     x_s = \begin{cases}
             1 & \text{if $s\in T$,}\\
             0 & \text{if there is no path from $s$ to $T$,} \\
             \sum_{s'\in\successors(s)}\prob(s,s')\cdot x_{s'} & \text{otherwise.}
           \end{cases}
  \]
\end{definition}

\begin{definition}[Expected discounted rewards]
  \label{def:rewards}
  For a discount factor $\beta\in (0,1)$ and a DTMC $\dtmc=(S,\sinit,\prob)$, the \emph{expected discounted reward} of a state state $s$
  is the unique solution of the following linear equation system:
  \begin{equation*}
    r_{s} = \rew(s) + \beta\cdot\!\!\!\!\!\!\sum_{s' \in \successors(s)}\!\!\!\!\prob(s,s')\cdot r_{s'}\text{\qquad for each $s\in S$.}
  \end{equation*}
\end{definition}
Recall that the problem to determine a policy that optimizes expected rewards
or probabilities is undecidable~\cite{MadaniHC99} in general.

\section{Solving POMDPs as MILPs}
\label{sec:milp}

While several sophisticated algorithms exist to compute policies for POMDPs, a simple, small, and
easy-to-analyze policy can be obtained by encoding the POMDP into a
\emph{Mixed Integer Linear Program} (MILP), which can be solved with linear optimization
tools like \emph{Gurobi}~\cite{gurobi}.
As a central advantage of the MILP formulations, it is straightforward to support multiple
specifications simultaniously. For instance, one can maximize the discounted reward
under the condition that the probability of reaching a target state is above a given bound
and the discounted cost below another threshold.

\subsection{Maximum Reachability Probabilities}
\label{subsec:mrp-milp}
Let $\pomdp = (\mdp,\obss,\obs)$ be a POMDP and $T\subseteq S$ a set of target states.
We assume that the states in $T$ have been made absorbing and that $\mdp$ contains only states from
which $T$ is reachable under at least one possible policy.
All other states can be removed from the POMDP.
We define the following MILP:
{\allowdisplaybreaks
\begin{subequations}
  \begin{align}
    \text{maximize}{ \colon }\quad & p_{\sinit} \label{eq:stat_sched:obj} \\
    \text{subject to}{ \colon } \quad & \notag \\
    \forall s\in\states\setminus T\colon \quad & \sum\limits_{\act\in\acts(s)} \sigma_{\obs(s),\act} = 1 \label{eq:stat_sched:sched} \\
    \forall s\in T\colon\quad & p_s = 1 \label{eq:stat_sched:target} \\
    \forall s\in\states\setminus T\ \forall\act\in\acts(s)\colon \quad &
       p_s\leq (1-\sigma_{\obs(s),\act}) + \!\!\!\!\sum_{s'\in\successors(s,\act)}\!\!\!\! \prob(s,\act,s')\cdot p_{s'}
       \label{eq:stat_sched:prob}
         \end{align}
\end{subequations}
\begin{subequations}
\begin{align}
    \forall (s,\act)\in\acts^{\mathrm{pr}}\ \forall s'\in\successors(s,\act)\colon\quad &
       r_s < r_{s'} + 1 - t_{s,s'} \label{eq:stat_sched:reach1} \\
    \forall (s,\act)\in\acts^{\mathrm{pr}}\colon\quad &
       p_s \leq 1 - \sigma_{\obs(s),\act} + \!\!\!\!\sum_{s'\in\successors(s,\act)}\!\!\!\! t_{s,s'}
       \label{eq:stat_sched:reach2}
  \end{align}
\end{subequations}}
The variables $p_s\in [0,1]$ store the probability to reach a target state from $s$ under the chosen policy.
We maximize this probability for the initial state $\sinit$ \eqref{eq:stat_sched:obj}. The variables $\sigma_{z,\act}$
for $z\in\obss$ and $\act\in\acts$ encode the policy.
$\sigma_{\obs(s),\act} = 1$ implies that the policy chooses action $\act$ in all states with observation $\obs(s)$ -- as we are computing stationary deterministic policies,
$\sigma_{\obs(s),\act} \in \{0,1\}$ for all $s\in S$ and $\act\in\acts(s)$. Thus, \eqref{eq:stat_sched:sched} ensures that for each observation exactly one action is selected.
\eqref{eq:stat_sched:target} ensures that target states are assigned a probability of $1$.
For non-target states $s\in\states\setminus T$, \eqref{eq:stat_sched:prob} recursively defines the probability $p_s$: for actions that are not chosen,
\ie $\sigma_{\obs(s),\act} = 0$, the inequality is always satisfied, as it can be simplified to $p_s \leq 1 + \epsilon$ with $\epsilon \geq 0$.
If $\sigma_{\obs(s),\act} = 1$, the probability is defined as the sum of the probability in each of the successors of the current state, multiplied with the probability to proceed to this successor when taking the current action. Maximizing the value of $p_s$ ensures that this constraint is satisfied by equality.
If the target states are reachable from all states under all possible policies, \eqref{eq:stat_sched:obj}--\eqref{eq:stat_sched:prob} are sufficient. We add \eqref{eq:stat_sched:reach1} and \eqref{eq:stat_sched:reach2} to avoid computing invalid values under policies that make the targets unreachable from some states: we define the problematic states $\states^{\mathrm{pr}}$ as the set of states that can only reach the target states under some policies, and compute them using standard graph algorithms. The problematic actions are then given by $\acts^{\mathrm{pr}} = \bigl\{ (s,\act)\in\states\times\acts\,\big|\,\act\in\acts(s)\land\successors(s,\act)\subseteq\states^{\mathrm{pr}}\bigr\}$.
We then introduce a ranking over the problematic states: each $s \in \states^{\mathrm{pr}}$ is assigned a value $r_s\in [0,1]$.
Next, we try to assign a transition to a successor state of $s$ by setting $t_{s,s'} = 1$ such that the value of the rank increases along the transition, \ie $r_{s'} > r_{s}$.
If this is not the case, \eqref{eq:stat_sched:reach1} enforces $t_{s,s'} = 0$. If the target state cannot be reached under the current policy, \ie $t_{s,s'} = 0$ for all successors of $s$, \eqref{eq:stat_sched:reach2} ensures that $p_s = 0$.
This technique is inspired by the reachability constraints from~\cite{wimmer-et-al-tcs-2014} that are used to compute counterexamples for MDPs~\cite{DBLP:conf/atva/DehnertJWAK14}.
An alternative formulation of reachability constraints using flow constraints can be found in \cite{Velasquez19}.

\subsection{Maximum Expected Discounted Rewards}
\label{subsec:medr-milp}
Let $\pomdp = (\mdp,\obss,\obs)$ be a POMDP. For a discount factor $\beta\in(0,1)$ and an upper bound on the
maximum discounted expected reward $v^\ast_{\max}$, we can built the MILP as follows:
\begin{subequations}
  \begin{align}
    \text{maximize}\colon\quad     & v_{\sinit} \\
    \text{subject to}\colon\quad   & \notag \\
    \forall s\in\states\colon\quad & \sum\limits_{\act\in\acts(s)} \sigma_{\obs(s),\act} = 1 \label{eq:discounted_reward:sched}\\
    \forall s\in\states\ \forall\act\in\acts(s)\colon\quad & v_s \leq v^\ast_{\max}\cdot (1-\sigma_{\obs(s),\act}) + r(s,\act) \notag \\*
        & \qquad {} + \beta\cdot\!\!\!\!\!\!\sum_{s'\in\successors(s,\act)}\!\!\!\prob(s,\act,s')\cdot v_{s'}
       \label{eq:discounted_reward:value}
  \end{align}
\end{subequations}

The MILP for maximum discounted reward is analogous to the formulation for maximum reachability, with the following differences:
The real-valued variables $v_s\in\mathbb{R}$ for each $s\in\states$ store the maximum discounted expected reward corresponding to the selected policy.

As $v_s$ can have values $> 1$, in \eqref{eq:discounted_reward:value}, we need an upper bound $v_{\max}^{*}$ on the maximum expected reward.
One possibility is setting $v_{\max}^{*}$ to the maximum expected discounted reward of the underlying MDP $\mdp$, which serves as an upper bound on the reward that can be achieved in $\pomdp$.
An alternative is using $v_{\max}^{\ast} := \frac1{1-\beta}\cdot \max_{s,\act}\rew(s,\act)$.
Since we no longer have any target states, the expected reward is computed for an infinite run of $\pomdp$ under the selected policy. $0 < \beta < 1$ guarantees that the expected reward converges to a finite number.
Thus, we don't need the reachability constraints we introduced in Sect.~\ref{subsec:mrp-milp}.
This simplification makes the MILP considerably smaller and more efficient to solve.

\subsection{Randomization}
\label{subsec:rand}

\begin{figure}[b]
  \centering
  \begin{tikzpicture}[bobbel/.style={circle,minimum size=1mm,inner sep=0mm,fill=black},>=stealth, scale=0.8]
    \begin{scope}
    \node[state] (s1) at (0,0) {$s_1$};
    \node[state,fill=yellow!20] (s2) at (2,1) {$s_2$};
    \node[state,fill=yellow!20] (s3) at (2,-1) {$s_3$};
    \node[state,fill=blue!20] (s4) at (4,-0) {$s_4$};

    \node[bobbel] at (1,0) {};

    \draw[->] (-1,0) -- node[above]{$\alpha$} (s1);
    \draw[-] (s1) -- node[above]{$\alpha$} (1,0);
    \draw[->] (1,0) -- node[above left]{$\frac{1}{2}$} (s2);
    \draw[->] (1,0) -- node[below left]{$\frac{1}{2}$} (s3);

    \draw[->] (s2) -- node[above]{$\alpha$} (s4);
    \draw[->] (s3) -- node[below]{$\beta$} (s4);

    \path
    (s2) edge [loop below] node[xshift=0.35cm, yshift=0.45cm] {$\beta$} (s2)
    (s3) edge [loop above] node[xshift=0.35cm, yshift=-0.45cm] {$\alpha$} (s3)
    (s4) edge [loop right] node {$\alpha$} (s4);
  \end{scope}
  \end{tikzpicture}
  \caption{A simple example that needs arbitrary randomization for maximum reachability of $s_4$.}
  \label{fig:randomization_needed}
\end{figure}
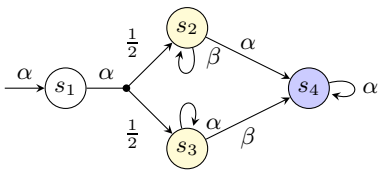
Stationary, deterministic policies can be restrictive in many use cases. 
However, while randomization
might often be necessary, sometimes the actual probability distribution does not matter. In
Fig.~\ref{fig:randomization_needed}, any stationary deterministic policy can reach the blue state with
probability of at most $0.5$. However, assigning any distribution with $\sigma_{\text{yellow},\alpha} > 0$ and
$\sigma_{\text{yellow},\beta} > 0$ leads to a probability of $1$.

In order to achieve this effect, we allow (besides deterministic choices) a randomized choice
with an arbitrary, but fixed distribution over the enabled actions.
This can be done by introducing an additional action $u\not\in\acts$ that is enabled in the 
set $\states'\subseteq S$ of states with (1) a non-unique observation and (2) more than one enabled action.
We replace the underlying MDP $\mdp$ by
$\mdp'=(\states,\sinit,\acts_u,\prob_u)$ such that $\acts_u\colonequals\acts\cup\{u\}$
and $\prob_u$ as follows: $\prob_u$ coincides with $\prob$ in states
$\states\setminus\states'$ and whenever $\act\neq u$. For instance, consider a state $s\in\states'$ where we want to achieve a uniform
distribution over all actions.
We set $\prob_u(s,u,s') = \frac{1}{|\acts(s)|}\cdot\sum_{\act\in\acts(s)}\prob(s,\act,s')$
for $s'\in\states$.

\begin{figure}[tb]
	\centering
	\subfloat[Pure]{
		\label{fig:a:gridpolicydet}
		\scalebox{0.6}{
			  \begin{tikzpicture}[bobbel/.style={circle,minimum size=1mm,inner sep=0mm,fill=black},>=stealth,font=\large]
    \begin{scope}
    \node[state,minimum size=1cm] (s0) at (0,0) {};
    \node[state,fill=blue!20,minimum size=1cm] (s2) at (4,0) {};
    \node[state,fill=yellow!20,minimum size=1cm] (s3) at (4,-2) {};
    \node[state,fill=green!20,minimum size=1cm] (s1) at (4,2) {};

    \draw[->] (s0) -- node[above]{$\alpha$} (s1);
    \draw[->] (s0) -- node[above]{$\beta$} (s2);
    \draw[->] (s0) -- node[above]{$\gamma$} (s3);
  \end{scope}
  \end{tikzpicture}
		}
	}
        \hspace{\fill}
	\subfloat[Light]{
		\label{fig:a:gridpolicyr}
		\scalebox{0.6}{
			  \begin{tikzpicture}[bobbel/.style={circle,minimum size=1mm,inner sep=0mm,fill=black},>=stealth,font=\large]
    \begin{scope}
    \node[state,minimum size=1cm] (s0) at (-2mm,0) {};
    \node[state,fill=blue!20,minimum size=1cm] (s2) at (4,0) {};
    \node[state,fill=yellow!20,minimum size=1cm] (s3) at (4,-2) {};
    \node[state,fill=green!20,minimum size=1cm] (s1) at (4,2) {};

    \node[bobbel] at (2,0) {};

    \draw[-, dashed] (s0) -- node[below left]{$\alpha$} (-1,1);
    \draw[-, dashed] (s0) -- node[right]{$\beta$} (0,1.3);
    \draw[-, dashed] (s0) -- node[below right]{$\gamma$} (1,1);
    \draw[-] (s0) -- node[below]{$\alpha + \beta + \gamma$} (2,0);
    \draw[->] (2,0) -- node[above]{$\nicefrac{1}{3}$} (s1);
    \draw[->] (2,0) -- node[above]{$\nicefrac{1}{3}$} (s2);
    \draw[->] (2,0) -- node[above]{$\nicefrac{1}{3}$} (s3);


  \end{scope}
  \end{tikzpicture}
		}
	}
        \hspace{\fill}
    \subfloat[Heavy]{
		\label{fig:a:gridpolicys}
		\scalebox{0.6}{
			  \begin{tikzpicture}[bobbel/.style={circle,minimum size=1mm,inner sep=0mm,fill=black},>=stealth,font=\large]
    \begin{scope}
    \node[state,minimum size=1cm] (s0) at (0,0) {};
    \node[state,fill=blue!20,minimum size=1cm] (s2) at (4,0) {};
    \node[state,fill=yellow!20,minimum size=1cm] (s3) at (4,-2) {};
    \node[state,fill=green!20,minimum size=1cm] (s1) at (4,2) {};

    \node[bobbel] at (2,0) {};
    \node[bobbel] at (2,-1) {};
    \node[bobbel] at (2,1) {};

    \draw[-, dashed] (s0) -- node[below]{$\alpha$} (-1.3,0);
    \draw[-, dashed] (s0) -- node[right]{$\beta$} (0,1.3);
    \draw[-, dashed] (s0) -- node[above]{$\gamma$} (-1,1);
    \draw[-, dashed] (s0) -- node[below]{$\alpha + \beta + \gamma$} (-1,-1);
    \draw[-] (s0) -- node[above, yshift=0.1cm]{$\alpha + \beta$} (2,1);
    \draw[-] (s0) -- node[above, yshift=-0.1cm]{$\alpha + \gamma$} (2,0);
    \draw[-] (s0) -- node[below, yshift=-0.1cm]{$\beta + \gamma$} (2,-1);
    \draw[->] (2,1) -- node[above] {$\nicefrac{1}{2}$} (s1);
    \draw[->] (2,1) -- node[below, xshift=0.1cm, yshift=0.0cm] {$\nicefrac{1}{2}$} (s2);
    
    \draw[->] (2,0) -- node[below, xshift=0.55cm, yshift=0.5cm] {$\nicefrac{1}{2}$} (s1);
    \draw[->] (2,0) -- node[above, xshift=0.5cm, yshift=-0.5cm] {$\nicefrac{1}{2}$} (s3);
    
    \draw[->] (2,-1) -- node[above, xshift=0.1cm, yshift=0.0cm] {$\nicefrac{1}{2}$} (s2);
    \draw[->] (2,-1) -- node[below]{$\nicefrac{1}{2}$} (s3);


  \end{scope}
  \end{tikzpicture}
		}
	}
	\caption{A POMDP without, with light, and with heavy static randomization}
	\label{fig:RandModes}
\end{figure}
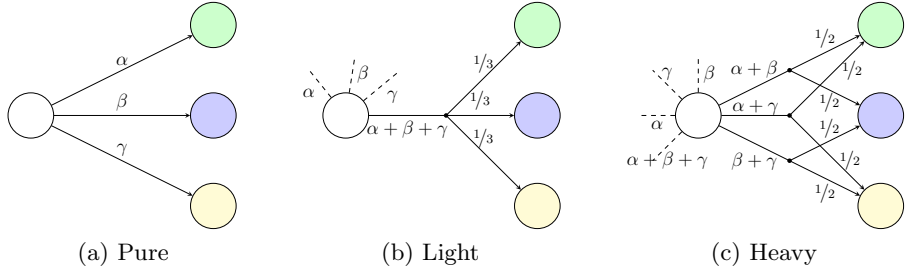

Any finite set of distributions can be supported that way.
We suggest three modes of randomization, as illustrated in Fig.~\ref{fig:RandModes}: \emph{Pure} (no randomization), \emph{Light} (adding one uniform distribution the enabled actions for each state) and \emph{Heavy} (adding a uniform distribution for each non-empty subset of enabled actions.

\section{Splitting Observations and States}
\label{sec:preprocessing}

Finding an optimal stationary policy
is a much easier problem than optimizing over all (history-dependent) policies, but the quality of stationary policies can be arbitrarily
worse than the quality of a general optimal policy. We attempt to preprocess POMDPs in a way that implicitly adds history 
locally by encoding previous observations into the states -- thus, making stationary policies computed on the augmented POMDP more powerful. 
In order to do so, we introduce \emph{observation splitting} and \emph{state splitting}.

\subsection{Observation Splitting}
\label{ssec:observation_splitting}

Let $\pomdp = (\mdp,\obss,\obs)$ be a POMDP with underlying MDP $\mdp=(\states,\sinit,\acts,\prob,\rew)$,
$z\in\obss$ an observation, and $\obs^{-1}(z)=\{s\in\states\,|\,z=\obs(s)\}$ the set of states
with observation $z$. W.\,l.\,o.\,g., let $\lvert\obs^{-1}(\obs(\sinit))\rvert = 1$ and $\prob(s,\sinit) = 0$
for all $s\in\states$. An existing POMDP can easily be modified to conform with these requirements.

\begin{definition}[Pre-Observations]
  \label{def:preobs}
  For $s\in\states$, the \textit{pre-observations} of $s$ are defined as
  $\text{pred}_\pomdp(s) = \bigl\{ (z,\act)\in\obss\times\acts\,\big|\,\exists s'\in\states\colon z=\obs(s')\land\prob(s',\act,s)>0\bigr\}$.
\end{definition}

Assume that $s,s'$ are the only states with observation $z=\lambda(s)=\lambda(s')$ and
that the pre-observations of $s$ are disjoint from the pre-observations of $s'$.
A history-dependent policy can easily distinguish the two states by remembering the previous observation and action, but a stationary
policy can not.
Observation splitting assigns distinct observations to the two states. While a memoryless policy on the original POMDP
has to make the same decision in $s$ and $s'$, it can make different decisions on the modified system. Therefore,
a memoryless policy on the modified system typically corresponds to a history-dependent policy on the original POMDP.
Note that this operation does not increase the number of states or transitions.

An observation $z$ can be split if we can partition $\obs^{-1}(z)$ into two disjoint subsets $A$ and $B$
such that $\left(\bigcup_{s\in A}\text{pred}_\pomdp(s)\right)\cap\left(\bigcup_{s\in B}\text{pred}_\pomdp(s)\right)=\emptyset$,
\ie when $z$ is observed, we
can distinguish states in $A$ from states in $B$ if the observation in the predecessor state as well as the last chosen action are known.
This information can be encoded into the POMDP by assigning distinct observations to the states in $A$ and the states in $B$.
Formally, we get the POMDP $\pomdp' = (\mdp,\obss',\obs')$ with 
\begin{itemize}
 \item[] $\obss' = (\obss\setminus\{z\})\dcup \{z_A,z_B\}$ and 
 $\obs'(s) =
     \begin{cases}
        \obs(s) & \text{if $s\notin A\dcup B$,} \\
        z_A     & \text{if $s\in A$,} \\
        z_B     & \text{if $s\in B$.}
     \end{cases}$
\end{itemize}

\begin{theorem}
Let $\pomdp'$ be the POMDP we obtain by splitting some observation $z$ of POMDP $\pomdp$ into new observations $z_A$ and $z_B$. Then:
  \begin{enumerate}
  \item $\{ \pomdp_\sched\,|\,\sched\in\schedset_\pomdp\} = \{\pomdp'_\sched\,|\,\sched\in\schedset_{\pomdp'}\}$, and
  \item $\{ \pomdp_\sched\,|\,\sched\in\schedset_\pomdp^{\text{stat}}\} \subseteq \{ \pomdp'_\sched\,|\,\sched\in\schedset_{\pomdp'}^{\text{stat}}\}$.
  \end{enumerate}
  \label{th_obs}
\end{theorem}
If we consider the set of all policies, observation splitting does not make a difference as we can obtain the same
induced DTMCs before and after observation splitting. However, if we only consider stationary policies, we get more
freedom and can choose among a larger set of induced DTMCs.

\begin{proof}
  Let $\pomdp$ be a POMDP and $\pomdp'$ result from $\pomdp$ by splitting observation $z$.
  Let, for $i=1,2$, $\pi_i = s_0^i\act_0^i s_1^i\act_1^i\ldots s_n^i\in\pathsfin$ be two
  finite paths in $\pomdp$, and $\pi_i'$ be the corresponding paths in $\pomdp'$. It is easy to see
  that $\lambda'(\pi'_1) = \lambda'(\pi'_2)$ iff $\lambda(\pi_1) = \lambda(\pi_2)$.
  That means, for each policy in $\pomdp$ there is a corresponding policy in $\pomdp'$ that
  makes the same decisions and vice versa.

  Additionally, for all states $s_1,s_2$ of $\pomdp$ and $\pomdp'$, we have
  $\lambda(s_1)\neq\lambda(s_2)\ \Rightarrow\ \lambda'(s_1)\neq\lambda'(s_2)$.
  Therefore a stationary policy that can make different choices in $s_1$ and $s_2$ in
  $\pomdp$ can make different choices in $\pomdp'$ as well.\qed
\end{proof}

\subsection{State Splitting}
\label{ssec:state_splitting}

\begin{figure}[tb]
  \centering
  \begin{tikzpicture}[bobbel/.style={circle,minimum size=1mm,inner sep=0mm,fill=black},>=stealth,
        state/.append style={minimum size=6.5mm},scale=0.9]
    \begin{scope}
    \node[state] (s0) at (-2,-1.5) {$s$};
    \node[state,fill=yellow!20] (s1) at (0,-0.5) {$s_1$};
    \node[state,fill=yellow!20] (s2) at (0,-1.5) {$s_2$};
    \node[state,fill=yellow!20] (s3) at (0,-2.5) {$s_3$};

    \node[bobbel] at (-1,-1) {};
    \node[bobbel] at (-1,-2) {};

    \draw[-] (s0) -- node[above left]{$\alpha$} (-1,-1);
    \draw[->] (-1,-1) -- node[above left]{$\nicefrac{1}{2}$} (s1);
    \draw[->] (-1,-1) -- node[above right,yshift=-1.5mm]{$\nicefrac{1}{2}$} (s2);

    \draw[-] (s0) -- node[below left]{$\beta$} (-1,-2);
    \draw[->] (-1,-2) -- node[below right,yshift=1.5mm]{$\nicefrac{1}{2}$} (s2);
    \draw[->] (-1,-2) -- node[below left]{$\nicefrac{1}{2}$} (s3);
  \end{scope}

  \begin{scope}[xshift=1cm,yshift=-1.5cm]
    \draw[->,thick] (0,0)-- node[above]{\footnotesize state} node[below]{\footnotesize splitting} (1,0);
  \end{scope}

  \begin{scope}[xshift=5cm,yshift=1cm]
    \node[state] (s0) at (-2,-2.5) {$s$};
    \node[state,fill=yellow!20] (s1) at (0,-1) {$s_1$};
    \node[state,fill=yellow!20] (s2a) at (0,-2) {$s_2^1$};
    \node[state,fill=yellow!20] (s2b) at (0,-3) {$s_2^2$};
    \node[state,fill=yellow!20] (s3) at (0,-4) {$s_3$};

    \node[bobbel] at (-1,-1.5) {};
    \node[bobbel] at (-1,-3.5) {};

    \draw[-] (s0) -- node[above left]{$\alpha$} (-1,-1.5);
    \draw[->] (-1,-1.5) -- node[above left]{$\nicefrac{1}{2}$} (s1);
    \draw[->] (-1,-1.5) -- node[below left]{$\nicefrac{1}{2}$} (s2a);

    \draw[-] (s0) -- node[below left]{$\beta$} (-1,-3.5);
    \draw[->] (-1,-3.5) -- node[above left]{$\nicefrac{1}{2}$} (s2b);
    \draw[->] (-1,-3.5) -- node[below left]{$\nicefrac{1}{2}$} (s3);
  \end{scope}

  \begin{scope}[xshift=6cm,yshift=-1.5cm]
    \draw[->,thick] (0,0)-- node[above]{\footnotesize observation} node[below]{\footnotesize splitting} (1,0);
  \end{scope}

  \begin{scope}[xshift=10cm,yshift=1cm]
    \node[state] (s0) at (-2,-2.5) {$s$};
    \node[state,fill=yellow!20] (s1) at (0,-1) {$s_1$};
    \node[state,fill=yellow!20] (s2a) at (0,-2) {$s_2^1$};
    \node[state,fill=blue!20] (s2b) at (0,-3) {$s_2^2$};
    \node[state,fill=blue!20] (s3) at (0,-4) {$s_3$};

    \node[bobbel] at (-1,-1.5) {};
    \node[bobbel] at (-1,-3.5) {};

    \draw[-] (s0) -- node[above left]{$\alpha$} (-1,-1.5);
    \draw[->] (-1,-1.5) -- node[above left]{$\nicefrac{1}{2}$} (s1);
    \draw[->] (-1,-1.5) -- node[below left]{$\nicefrac{1}{2}$} (s2a);

    \draw[-] (s0) -- node[below left]{$\beta$} (-1,-3.5);
    \draw[->] (-1,-3.5) -- node[above left]{$\nicefrac{1}{2}$} (s2b);
    \draw[->] (-1,-3.5) -- node[below left]{$\nicefrac{1}{2}$} (s3);
  \end{scope}
  \end{tikzpicture}
  \caption{Applying state splitting and observation splitting to a POMDP. The observations are
    given by the color of the states.}
  \label{fig:state_and_observation_splitting}
\end{figure}
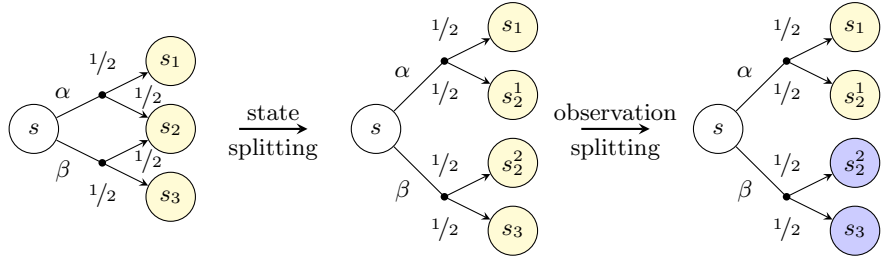

Often, observation splitting is not applicable to a given POMDP.
We define state splitting for refining the POMDP to enable observation splitting:
In Fig.~\ref{fig:state_and_observation_splitting}, all states that have the same color share an observation (\ie $\obs(s_1)=\obs(s_2)=\obs(s_3)$).
We have $\text{pred}_\pomdp(s_2) = \text{pred}_\pomdp(s_1) \cup \text{pred}_\pomdp(s_3)$, so the three states cannot be split into disjoint sets by means of their pre-observations and thus, observation splitting cannot be applied.
However, by creating two copies $s_2^1$ and $s_2^2$, the pre-observations of $s_2^1$ and $s_1$ on the one hand and $s_2^2$ and $s_3$ on the other hand become disjoint, thus enabling observation splitting on the yellow observation.

Formally, we can split a state $s\in\states$ in a POMDP $\pomdp=(\mdp,\obss,\obs)$ whenever $|\text{pred}_\pomdp(s)|>1$.
Again, we assume that $\text{pred}_\pomdp(\sinit) = \emptyset$. We obtain a modified POMDP $\pomdp' = (\mdp',\obss',\obs')$ with 
\begin{itemize}
        \item $\states' \coloneqq \bigl(\states\setminus\{s\}\bigr)\dcup\bigl\{ (s,z,\act)\,\big|\,(z,\act)\in\text{pred}_\pomdp(s)\bigr\}$;\\[-0.5\baselineskip]
        \item for all $t\in\states'$ we set:
            $\obs'(t)\coloneqq\obs(s)$ if $t=(s,z,\act)$ for some $z\in\obss$ and $\act\in\acts$,
            and $\obs'(t)\coloneqq\obs(t)$ otherwise;
        \item for all $t,t'\in\states'$, $\beta\in\acts$:
            \[
            \prob'(t,\beta,t') \coloneqq \begin{cases}
                    \prob(t,\beta,t') & \text{if $t,t'\in\states\setminus\{s\}$,} \\
                    \prob(s,\beta,s)  & \text{if $t=(s,z,\act)$ and $t' = (s,z,\beta)$} \\
                                      & \text{\qquad for some $z\in\obss$ and $\act\in\acts$,} \\
                    \prob(t,\beta,s)  & \text{if $t\in\states\setminus\{s\}$ and $t'=(s,\obs(t),\beta)$,} \\
                    \prob(s,\beta,t') & \text{if $t=(s,z,\act)$ for some $z\in\obss$} \\
                                      & \text{\qquad and $\act\in\acts$ and $t'\in\states\setminus\{s\}$,}\\
                    0                 & \text{otherwise;}
                \end{cases}
            \]
        \item for all $t\in\states'$ and $\beta\in\acts(t)$ we set:
            $\rew'(t,\beta) \coloneqq \rew(s,\beta)$ if $t=(s,z,\act)$ for some $z\in\obss$ and $\act\in\acts$,
            and $\rew'(t,\beta)\coloneqq\rew(t,\beta)$ otherwise.
    \end{itemize}

\begin{theorem}
  Let $\pomdp$ be a POMDP and let $\pomdp'$ result from $\pomdp$ by splitting a state $s$ of $\pomdp$. Then
  $\pomdp$ and $\pomdp'$ are bisimilar.
  \label{th_state}
\end{theorem}
After extending the definition of bisimulation to POMDPs, this can be proven by defining an equivalence relation between $s$ and the states $(s,z,\act)$ produced by splitting it.

It is well known~\cite{GivanDG03} that bisimilar systems satisfy (among others)
the same LTL and PCTL properties, including reachability
and discounted expected rewards.

\section{Implementation}
\label{sec:implementation}

We implemented both MILP formulations described in Sect.~\ref{sec:milp} and use the commercial solver Gurobi~\cite{gurobi} to solve them.
From our experience, Gurobi often finds a feasible solution, which satisfies all constraints, quickly, but then spends a lot of time trying to improve this initial solution or prove its optimality.
However, even this initial solution is often already close to the optimum.
Thus, we have implemented a \emph{time limit} mode, in which the solver tries to optimize the result for a predefined number of seconds after the first solution is found.

We implemented the MILPs with three different levels of randomization as in Sect.~\ref{subsec:rand}, and observation and state splitting as in Sect.~\ref{sec:preprocessing}.

State splitting by itself only increases the size of the state space and yields a bisimilar system. Therefore
it only makes sense to apply state splitting when it enables observation splitting, which in turn
increases the power of stationary policies.
However, it is not clear beforehand which states to split.
So as a rule of thumb, we want to determine a small subset of states whose splitting enables a large number of observation splits.

\paragraph{Splitting Heuristic.}

We suggest a splitting heuristic that uses previous results of the MILP to iteratively refine the POMDP by selecting states for splitting,
see Algorithm~\ref{algo:splitting} for an outline.
\begin{algorithm}[tb]
   \small
   \caption{Splitting Heuristic}
   \label{algo:splitting}
   \KwIn{POMDP $\pomdp$}
   oldResult $\gets$ 0\;
   splitObservations($\pomdp$)\;
   newResult $\gets$ computeMILP($\pomdp$)\;
   \While{newResult $>$ oldResult}{
      oldResult $\gets$ newResult\;
      splitGroup $\gets$ computeSplitGroup($\pomdp$, oldResult)\;
      splitStates($\pomdp$, splitGroup)\;
      splitObservations($\pomdp$)\;
      newResult $\gets$ computeMILP($\pomdp$)\;
   }
\end{algorithm}
First, we apply observation splitting on the original POMDP and compute the optimal stationary policy in that POMDP to get a baseline for the following optimization.
Then, we use this solution to determine a set of states for splitting. 
Similar to what policy iteration for MDPs~\cite{puterman} does, we check if locally changing a selected action would result in an improvement. $\sched^\ast$ is the current policy and
$v^\ast_s$ the corresponding value of state $s$. We choose
\[
\sigma'(s):\in\argmax_{\act\in\acts(s)}\sum_{s'\in\states}\prob(s,\act,s')\cdot v^\ast_{s'},
\]
preferring $\sigma'(s)=\sigma^\ast(s)$ where possible.

Whenever $\sigma'(s)\neq\sigma^\ast(s)$ holds, then being able to distinguish $s$ from the other states with the same observation would lead to an improvement.
Therefore $s$ is added to the set \emph{splitGroup}.
State splitting is applied to all states in \emph{splitGroup}.
Afterwards we apply observation splitting as long as it modifies the POMDP, and solve the MILP for the modified POMDP.
We repeat this procedure, until no further improvements can be made.

In case of multiple specifications, it can happen that the initial MILP is infeasible on the original POMDP.
In this case we apply Algorithm~\ref{algo:splitting} to optimize the first constraint until it is satisfied.
Then we optimize the second one under the condition that the first constraint is satisfied, etc.
In the end, we either obtain a policy that satisfies all constraints, or at some point we cannot satisfy one of the specifications.
This can have two reasons: either the POMDP does not satisfy the specification or state plus observation splitting are not powerful enough to yield a POMDP on which a stationary policy satisfies the constraints.
Note that a complete method does not exist due to the undecidability of the problem.

\section{Experiments}
\label{sec:experiments}

\paragraph{Experimental Setup.}
All experiments were run on a machine with a 3.3~GHz
Intel\textsuperscript{\textregistered} Xeon\textsuperscript{\textregistered} E5-2643 CPU
and 64~GB RAM, running Ubuntu 16.04.

We consider seven benchmarks from two different sources. The \emph{4$\times$4grid\_avoid} was taken from the \prismpomdp model checker\footnote{\url{http://www.prismmodelchecker.org/files/rts-poptas/}} and is a maximum reachability probability grid world (with one absorbing ``bad'' state that needs to be avoided). The other benchmarks were adopted from the POMDP page\footnote{\url{http://www.pomdp.org/examples/}} and slightly modified to fit our definitions. \emph{1d}, \emph{4$\times$4.95}, \emph{cheese.95}, \emph{mini-hall2}, and \emph{parr95.95} are grid worlds in which a reward is issued for reaching certain states. \emph{shuttle.95} describes a space shuttle delivering cargo between two space stations, and a reward is issued for every successful delivery (see Fig.~\ref{fig:shuttle}).

For two of the benchmarks, we added secondary constraints to demonstrate the effectiveness of our approach to multi-objective model checking.
On 4$\times$4grid\_avoid, we added a cost of $1$ for each step in the grid (except for the self loops in the goal and bad state). We require the reachability probability to be at least $0.25$, and minimize the (un-discounted) expected reward. Note that computing un-discounted reward is sound in this case, as we asure the computation of a valid policy by the reachability contraints as seen in \ref{subsec:mrp-milp} and a sink state is eventually reached with probability $1$. On cheese.95, we added a new state -- each time the goal state is reached, there is a choice to continue back into the maze, or to transit to a rewardless sink state. We then declared one state of the grid ``bad'' and required that the probability to reach this state is at most $0.5$, while still maximizing the total expected discounted reward.

We run our MILP implementation using Gurobi~8.1 to solve all benchmarks. To improve runtimes, we used time limits of 5, 10, 30, and 60\,s for the optimization part of each solver call (see Sect.~\ref{sec:implementation}). We also let the optimization run to termination (with a total time limit of 7200\,s) to get an assessment of the quality of the solutions that were found.

For comparison, we also ran the maximum expected reward benchmarks with the explicit point based POMDP solvers \emph{SARSOP}~\cite{6284837} and \emph{solvePOMDP}\footnote{\url{https://www.erwinwalraven.nl/solvepomdp/}}. The results for the maximum reachability probability benchmark (4$\times$4grid\_avoid) were compared against \prismpomdp. All solvers were run using standard parameters.
We did not find any solver that could handle the type of multi-objective model checking we implemented for 4$\times$4.95 and cheese.95.

\setlength\tabcolsep{3pt}
\begin{table}[tb]
 \caption{Results for different benchmarks, timeouts, and implementations}
 \label{tab:results}
 \begin{adjustbox}{width=\columnwidth,center}
    \begin{tabular}{lrccccccccc}
      \toprule
      Benchmark         & TO  & Pure       & Light       & Heavy        & Pure + H    & Light + H    & Heavy + H & SARSOP &solvePOMDP & \prismpomdp \\
      \midrule
       1d               & 5\,s       & \textbf{0.61/0.1s}  & \textbf{0.65/0.1s}   & \textbf{0.65/0.1s}  &  \textbf{0.83/0.1s}  &  \textbf{0.83/0.7s}   &  \textbf{0.83/0.1s} &0.95/0.003s & 0.95/1.3s & ---\\
                        & 10\,s      &            &             &            &             &              & \\
                        & 30\,s      &            &             &            &             &              & \\
                        & 60\,s      &            &             &            &             &              & \\
      4$\times$4.95            & 5\,s       & \textbf{0.22/0.1s}  & \textbf{0.41/0.4s}    & \textbf{3.0/0.7s}     & \textbf{0.22/0.5s}  & 3.55/34.3s  & \textbf{3.0/4.0s} &3.55/0.05s & 3.55/20.5s & ---\\
                        & 10\,s      &            &             &            &   & 3.55/71.6s &\\
                        & 30\,s      &            &             &            &      & \textbf{3.55/209.2s} &\\
                        & 60\,s      &            &             &            &  & &\\
      4$\times$4grid\_avoid    & 5\,s       & \textbf{0.21/0.1s}  & \textbf{0.3/0.2s}    & \textbf{0.85/0.1s}  & \textbf{0.21/0.2s} & \textbf{0.88/3.3s} & \textbf{0.93/9.3s} &--- & --- &0.96/346.9s\\
                        & 10\,s      &            &             &            &             &              &\\
                        & 30\,s      &            &             &            &             &              &\\
                        & 60\,s      &            &             &            &             &              &\\
      4$\times$4grid\_avoid    & 5\,s       & \textbf{UNSAT/0.1s}  & \textbf{13.63/0.1s}    & \textbf{4.4/0.2s}  & \textbf{UNSAT/0.1s} & \textbf{3.43/2.8s} & 4.40/17.2s* &---&---&---\\
      (p$\geq$0.25, MinR)  & 10\,s &   &   &   &   &   & 4.40/34.8s*\\
                        & 30\,s &   &   &   &   &   & 4.21/112.3s*\\
                        & 60\,s &   &   &   &   &   & 3.95/456.5s*\\
      cheese.95         & 5\,s        & \textbf{0.62/0.6s}  & \textbf{1.2/1.6s}    & 1.84/19s*   & 3.31/35.3s & 1.2/16.7s &  1.84/34.7s*&3.40/0.03s & 3.40/13.7s & ---\\
                        & 10\,s       &            &             & 1.84/37.7s*     & 3.31/72.7s &  2.06/71.7s & 1.84/70.3s*\\
                        & 30\,s       &            &             & 1.84/113.7s*     & 3.34/226.2s* & 2.1/222.8s* & 1.84/217s*\\
                        & 60\,s       &            &             & \textbf{1.84/162.5s}     & 3.34/454.3s* & 2.1/452s*   & 1.84/382.4s*\\
      cheese.95         & 5\,s        & \textbf{0.40/0.8s}  & \textbf{0.45/1.9s}    &  0.47/19.1s*  & \textbf{0.40/0.8s} & 0.50/39.8s & 0.47/36.4s* & --- & --- & ---\\
      (p$\leq$0.5, MaxR)    & 10\,s &  &  & 0.47/37.5s*&  & 0.50/77.7s  & 0.47/73.3s*\\
                        & 30\,s &   &   & 0.47/116.2s*  &   & 0.50/232.2s & 0.47/229.3s*\\
                        & 60\,s &   &   & \textbf{0.47/148.7s} &   & 0.51/464.6s*  & 0.47/370.8s*\\
      mini-hall2        & 5\,s       & \textbf{2.43/0.4s}  & \textbf{2.43/12s}  & 2.43/18.1s & 2.5/20.1s & 2.43/29.5s* &2.43/33.6s &2.71/0.04s & 2.71/33.8s & ---\\
                        & 10\,s      &            &             & 2.43/37.2s & 2.58/38.2s*  & 2.43/46.3s* & 2.43/71.2s\\
                        & 30\,s      &            &             & 2.46/114.2s     &  2.43/114.2s& 2.43/121.5s* & 2.46/213s\\
                        & 60\,s      &            &             & \textbf{2.51/228s}  & 2.58/227.9s* & 2.43/235s* & 2.51/434.9s*\\
      parr95.95         & 5\,s       & \textbf{6.0/0.2s}   & \textbf{6.0/0.2s}    & \textbf{6.0/0.2s}   & \textbf{6.84/0.5s} &  \textbf{6.84/0.5s}  & \textbf{6.84/0.7s} &6.84/0.02s & 6.84/8.1s &---\\
                        & 10\,s      &            &             &            &             &              & \\
                        & 30\,s      &            &             &            &             &              &\\
                        & 60\,s      &            &             &            &             &              & \\
      shuttle.95        & 5\,s       & \textbf{18.0/0.2s} & \textbf{18.0/0.4s}  & \textbf{18.0/1.0s} & 31.25/36.8s*& 31.25/34.1s* & 18.63/18.3s&31.25/0.05s & 31.25/804s&---\\
                        & 10\,s      &            &             &            & 31.25/74.6s & 31.25/71.1s* & 22.8/67.3s\\
                        & 30\,s      &            &             &            & 31.25/226s* & 31.25/223.2s* & 31.25/217.3s*\\
                        & 60\,s      &            &             &            & 31.25/452s* & 31.25/451.5s* & 31.25/443.6s*    & &\\
      \bottomrule
    \end{tabular}
    \end{adjustbox}
\end{table}

\begin{figure}[t]
	\centering
	\subfloat[cheese.95]{
		\label{fig:cheeseplot}
		\scalebox{0.7}{
		\begin{tikzpicture}
 \begin{scope}
 
  
  \draw[->, line width=0.5mm] (0cm,0cm) -- (6cm,0cm) node [right] {t(s)};  
  \draw[->, line width=0.5mm] (0cm,0cm) -- (0cm,4cm) node [above] {$v_{\sinit}$};  
  
  \draw[line width=0.5mm, text=black] (1 cm, -0.1cm) -- (1cm, 0.1cm) 
    node at (1cm, -0.35cm) {$100$};  
  \draw[line width=0.5mm, text=black] (3 cm, -0.1cm) -- (3cm, 0.1cm) 
    node at (3cm, -0.35cm) {$300$};  
  \draw[line width=0.5mm, text=black] (5 cm, -0.1cm) -- (5cm, 0.1cm) 
    node at (5cm, -0.35cm) {$500$};  

  \draw (-0.2cm, -0.2cm) node {$0$};
  
  \draw[line width=0.5mm, text=black] (-0.1 cm, 1cm) -- (0.1cm, 1cm) 
    node at (-0.4 cm, 1cm) {$1$};  
  \draw[line width=0.5mm, text=black] (-0.1 cm, 2cm) -- (0.1cm, 2cm) 
    node at (-0.4 cm, 2cm) {$2$};  
  \draw[line width=0.5mm, text=black] (-0.1 cm, 3cm) -- (0.1cm, 3cm) 
    node at (-0.4 cm, 3cm) {$3$};  

\fill[blue] (0.01cm,0.62cm) circle (0.06cm);
\node[blue] at (-0.4, 0.5) {Pure};

\fill[blue] (0.02cm,1.2cm) circle (0.06cm);
\node[blue] at (-0.45, 1.3) {Light};

\fill[blue] (0.19cm,1.84cm) circle (0.06cm);
\fill[blue] (0.38cm,1.84cm) circle (0.06cm);
\fill[blue] (1.14cm,1.84cm) circle (0.06cm);
\fill[blue] (1.63cm,1.84cm) circle (0.06cm);
\draw[dashed, line width=0.25mm] (0.19cm,1.84cm) -- (0.38cm,1.84cm);
\draw[dashed, line width=0.25mm] (0.38cm,1.84cm) -- (1.14cm,1.84cm);
\draw[dashed, line width=0.25mm] (1.14cm,1.84cm) -- (1.63cm,1.84cm);
\node[blue] at (1.6, 1.6) {Heavy};

\fill[red] (0.35cm,3.31cm) circle (0.06cm);
\fill[red] (0.73cm,3.31cm) circle (0.06cm);
\fill[red] (2.24cm,3.34cm) circle (0.06cm);
\fill[red] (4.54cm,3.34cm) circle (0.06cm);
\draw[dashed, line width=0.25mm] (0.35cm,3.31cm) -- (0.73cm,3.31cm);
\draw[dashed, line width=0.25mm] (0.73cm,3.31cm) -- (2.24cm,3.34cm);
\draw[dashed, line width=0.25mm] (2.24cm,3.34cm) -- (4.54cm,3.34cm);
\node[red] at (4.5, 3.5) {Pure + H};

\fill[red] (0.17cm,1.2cm) circle (0.06cm);
\fill[red] (0.72cm,2.06cm) circle (0.06cm);
\fill[red] (2.23cm,2.1cm) circle (0.06cm);
\fill[red] (4.52cm,2.1cm) circle (0.06cm);
\draw[dashed, line width=0.25mm] (0.17cm,1.2cm) -- (0.72cm,2.06cm);
\draw[dashed, line width=0.25mm] (0.72cm,2.06cm) -- (2.23cm,2.1cm);
\draw[dashed, line width=0.25mm] (2.23cm,2.1cm) -- (4.52cm,2.1cm);
\node[red] at (4.5, 2.3) {Light + H};

\fill[red] (0.35cm,1.84cm) circle (0.06cm);
\fill[red] (0.70cm,1.84cm) circle (0.06cm);
\fill[red] (2.17cm,1.84cm) circle (0.06cm);
\fill[red] (3.82cm,1.84cm) circle (0.06cm);
\draw[dashed, line width=0.25mm] (0.35cm,1.84cm) -- (0.70cm,1.84cm);
\draw[dashed, line width=0.25mm] (0.70cm,1.84cm) -- (2.17cm,1.84cm);
\draw[dashed, line width=0.25mm] (2.17cm,1.84cm) -- (3.82cm,1.84cm);
\node[red] at (3.8, 1.6) {Heavy + H};

\end{scope}
\end{tikzpicture}
		}
	}
	\subfloat[4$\times$4.95]{
		\label{fig:grid95plot}
		\scalebox{0.7}{
		\begin{tikzpicture}
 \begin{scope}
 
  
  \draw[->, line width=0.5mm] (0cm,0cm) -- (7cm,0cm) node [right] {t(s)};  
  \draw[->, line width=0.5mm] (0cm,0cm) -- (0cm,4cm) node [above] {$v_{\sinit}$};  
  
  \draw[line width=0.5mm, text=black] (2 cm, -0.1cm) -- (2cm, 0.1cm) 
    node at (2cm, -0.35cm) {$100$};  
  \draw[line width=0.5mm, text=black] (4 cm, -0.1cm) -- (4cm, 0.1cm) 
    node at (4cm, -0.35cm) {$200$};  
  \draw[line width=0.5mm, text=black] (6 cm, -0.1cm) -- (6cm, 0.1cm) 
    node at (6cm, -0.35cm) {$300$};  

  \draw (-0.2cm, -0.2cm) node {$0$};
  
  \draw[line width=0.5mm, text=black] (-0.1 cm, 1cm) -- (0.1cm, 1cm) 
    node at (-0.4 cm, 1cm) {$1$};  
  \draw[line width=0.5mm, text=black] (-0.1 cm, 2cm) -- (0.1cm, 2cm) 
    node at (-0.4 cm, 2cm) {$2$};  
  \draw[line width=0.5mm, text=black] (-0.1 cm, 3cm) -- (0.1cm, 3cm) 
    node at (-0.4 cm, 3cm) {$3$};  

\fill[purple] (0.002cm,0.22cm) circle (0.06cm);
\node[blue] at (-0.5, 0.05) {Pure};

\fill[blue] (0.008cm,0.41cm) circle (0.06cm);
\node[blue] at (-0.5, 0.5) {Light};

\fill[blue] (0.03cm,3.0cm) circle (0.06cm);
\node[blue] at (-0.6, 2.8) {Heavy};

\fill[purple] (0.01cm,0.22cm) circle (0.06cm);
\node[red] at (0.8, 0.3) {Pure + H};

\fill[red] (0.68cm,3.55cm) circle (0.06cm);
\fill[red] (1.44cm,3.55cm) circle (0.06cm);
\fill[red] (4.18cm,3.55cm) circle (0.06cm);
\draw[dashed, line width=0.25mm] (0.68cm,3.55cm) -- (1.44cm,3.55cm);
\draw[dashed, line width=0.25mm] (1.44cm,3.55cm) -- (4.18cm,3.55cm);
\node[red] at (4.2, 3.8) {Light + H};

\fill[red] (0.16cm,3.0cm) circle (0.06cm);
\node[red] at (1.0, 3.2) {Heavy + H};

\end{scope}
\end{tikzpicture}
		}
	}
	\caption{Probability and runtime of the different MILP approaches for different grid world benchmarks}
	\label{fig:Plots}
\end{figure}
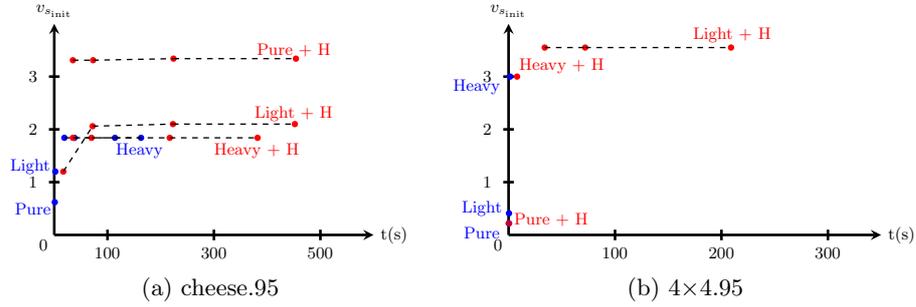

\begin{figure}[t]
	\centering
	\subfloat[Pure]{
		\label{fig:gridpolicydet}
		\scalebox{0.4}{
			\begin{tikzpicture}[shorten >=0.5mm,state/.style={circle, draw, minimum size=1cm}]]
\node[state] (s1) {$s_1$};
\node[state, right=1.5cm of s1] (s2) {$s_5$};
\node[state, right=1.5cm of s2] (s3) {$s_9$};
\node[state, fill=blue!20, right=1.5cm of s3] (s4) {$s_{13}$};

\node[state, above=1.5cm of s1] (s5) {$s_2$};
\node[state, right=1.5cm of s5] (s6) {$s_6$};
\node[state, right=1.5cm of s6] (s7) {$s_{10}$};
\node[state, fill=red!20, right=1.5cm of s7] (s8) {$s_{14}$};

\node[state, above=1.5cm of s5] (s9) {$s_3$};
\node[state, right=1.5cm of s9] (s10) {$s_7$};
\node[state, right=1.5cm of s10] (s11) {$s_{11}$};
\node[state, right=1.5cm of s11] (s12) {$s_{15}$};

\node[state, above=1.5cm of s9] (s13) {$s_4$};
\node[state, right=1.5cm of s13] (s14) {$s_8$};
\node[state, right=1.5cm of s14] (s15) {$s_{12}$};
\node[state, right=1.5cm of s15] (s16) {$s_{16}$};

\draw[->, line width = 2pt] (s1.south east) -- (s2.south west);
\draw[->] (s2.north west) -- (s1.north east);
\draw[->, line width = 2pt] (s2.south east) -- (s3.south west);
\draw[->] (s3.north west) -- (s2.north east);
\draw[->, line width = 2pt] (s3.south east) -- (s4.south west);

\draw[->, line width = 2pt] (s5.south east) -- (s6.south west);
\draw[->] (s6.north west) -- (s5.north east);
\draw[->, line width = 2pt] (s6.south east) -- (s7.south west);
\draw[->] (s7.north west) -- (s6.north east);
\draw[->, line width = 2pt] (s7.south east) -- (s8.south west);

\draw[->, line width = 2pt] (s9.south east) -- (s10.south west);
\draw[->] (s10.north west) -- (s9.north east);
\draw[->, line width = 2pt] (s10.south east) -- (s11.south west);
\draw[->] (s11.north west) -- (s10.north east);
\draw[->, line width = 2pt] (s11.south east) -- (s12.south west);
\draw[->] (s12.north west) -- (s11.north east);

\draw[->, line width = 2pt] (s13.south east) -- (s14.south west);
\draw[->] (s14.north west) -- (s13.north east);
\draw[->, line width = 2pt] (s14.south east) -- (s15.south west);
\draw[->] (s15.north west) -- (s14.north east);
\draw[->, line width = 2pt] (s15.south east) -- (s16.south west);
\draw[->] (s16.north west) -- (s15.north east);

\draw[->] (s5.south west) -- (s1.north west);
\draw[->] (s1.north east) -- (s5.south east);
\draw[->] (s9.south west) -- (s5.north west);
\draw[->] (s5.north east) -- (s9.south east);
\draw[->] (s13.south west) -- (s9.north west);
\draw[->] (s9.north east) -- (s13.south east);

\draw[->] (s6.south west) -- (s2.north west);
\draw[->] (s2.north east) -- (s6.south east);
\draw[->] (s10.south west) -- (s6.north west);
\draw[->] (s6.north east) -- (s10.south east);
\draw[->] (s14.south west) -- (s10.north west);
\draw[->] (s10.north east) -- (s14.south east);

\draw[->] (s7.south west) -- (s3.north west);
\draw[->] (s3.north east) -- (s7.south east);
\draw[->] (s11.south west) -- (s7.north west);
\draw[->] (s7.north east) -- (s11.south east);
\draw[->] (s15.south west) -- (s11.north west);
\draw[->] (s11.north east) -- (s15.south east);

\draw[->] (s12.south west) -- (s8.north west);
\draw[->] (s16.south west) -- (s12.north west);
\draw[->] (s12.north east) -- (s16.south east);

\path
    (s4) edge [loop above, line width = 2pt] (s4)
    (s8) edge [loop above, line width = 2pt] (s8);

\end{tikzpicture}
			\hspace*{0.5cm}
		}
	}
	\subfloat[Light]{
		\label{fig:gridpolicyr}
		\scalebox{0.4}{
			\hspace*{0.5cm}
			\begin{tikzpicture}[shorten >=0.5mm,state/.style={circle, draw, minimum size=1cm}]]
\node[state] (s1) {$s_1$};
\node[state, right=1.5cm of s1] (s2) {$s_5$};
\node[state, right=1.5cm of s2] (s3) {$s_9$};
\node[state, fill=blue!20, right=1.5cm of s3] (s4) {$s_{13}$};

\node[state, above=1.5cm of s1] (s5) {$s_2$};
\node[state, right=1.5cm of s5] (s6) {$s_6$};
\node[state, right=1.5cm of s6] (s7) {$s_{10}$};
\node[state, fill=red!20, right=1.5cm of s7] (s8) {$s_{14}$};

\node[state, above=1.5cm of s5] (s9) {$s_3$};
\node[state, right=1.5cm of s9] (s10) {$s_7$};
\node[state, right=1.5cm of s10] (s11) {$s_{11}$};
\node[state, right=1.5cm of s11] (s12) {$s_{15}$};

\node[state, above=1.5cm of s9] (s13) {$s_4$};
\node[state, right=1.5cm of s13] (s14) {$s_8$};
\node[state, right=1.5cm of s14] (s15) {$s_{12}$};
\node[state, right=1.5cm of s15] (s16) {$s_{16}$};

\draw[->, line width = 2pt] (s1.south east) -- (s2.south west);
\draw[->, line width = 2pt] (s2.north west) -- (s1.north east);
\draw[->, line width = 2pt] (s2.south east) -- (s3.south west);
\draw[->, line width = 2pt] (s3.north west) -- (s2.north east);
\draw[->, line width = 2pt] (s3.south east) -- (s4.south west);

\draw[->, line width = 2pt] (s5.south east) -- (s6.south west);
\draw[->, line width = 2pt] (s6.north west) -- (s5.north east);
\draw[->, line width = 2pt] (s6.south east) -- (s7.south west);
\draw[->, line width = 2pt] (s7.north west) -- (s6.north east);
\draw[->, line width = 2pt] (s7.south east) -- (s8.south west);

\draw[->, line width = 2pt] (s9.south east) -- (s10.south west);
\draw[->, line width = 2pt] (s10.north west) -- (s9.north east);
\draw[->, line width = 2pt] (s10.south east) -- (s11.south west);
\draw[->, line width = 2pt] (s11.north west) -- (s10.north east);
\draw[->, line width = 2pt] (s11.south east) -- (s12.south west);
\draw[->, line width = 2pt] (s12.north west) -- (s11.north east);

\draw[->, line width = 2pt] (s13.south east) -- (s14.south west);
\draw[->, line width = 2pt] (s14.north west) -- (s13.north east);
\draw[->, line width = 2pt] (s14.south east) -- (s15.south west);
\draw[->, line width = 2pt] (s15.north west) -- (s14.north east);
\draw[->, line width = 2pt] (s15.south east) -- (s16.south west);
\draw[->, line width = 2pt] (s16.north west) -- (s15.north east);

\draw[->, line width = 2pt] (s5.south west) -- (s1.north west);
\draw[->, line width = 2pt] (s1.north east) -- (s5.south east);
\draw[->, line width = 2pt] (s9.south west) -- (s5.north west);
\draw[->, line width = 2pt] (s5.north east) -- (s9.south east);
\draw[->, line width = 2pt] (s13.south west) -- (s9.north west);
\draw[->, line width = 2pt] (s9.north east) -- (s13.south east);

\draw[->, line width = 2pt] (s6.south west) -- (s2.north west);
\draw[->, line width = 2pt] (s2.north east) -- (s6.south east);
\draw[->, line width = 2pt] (s10.south west) -- (s6.north west);
\draw[->, line width = 2pt] (s6.north east) -- (s10.south east);
\draw[->, line width = 2pt] (s14.south west) -- (s10.north west);
\draw[->, line width = 2pt] (s10.north east) -- (s14.south east);

\draw[->, line width = 2pt] (s7.south west) -- (s3.north west);
\draw[->, line width = 2pt] (s3.north east) -- (s7.south east);
\draw[->, line width = 2pt] (s11.south west) -- (s7.north west);
\draw[->, line width = 2pt] (s7.north east) -- (s11.south east);
\draw[->, line width = 2pt] (s15.south west) -- (s11.north west);
\draw[->, line width = 2pt] (s11.north east) -- (s15.south east);

\draw[->, line width = 2pt] (s12.south west) -- (s8.north west);
\draw[->, line width = 2pt] (s16.south west) -- (s12.north west);
\draw[->,line width = 2pt] (s12.north east) -- (s16.south east);

\path
    (s4) edge [loop above, line width = 2pt] (s4)
    (s8) edge [loop above, line width = 2pt] (s8);

\end{tikzpicture}
			\hspace*{0.5cm}
		}
	}
    \subfloat[Heavy]{
		\label{fig:gridpolicys}
		\scalebox{0.4}{
			\hspace*{0.5cm}
			\begin{tikzpicture}[shorten >=0.5mm,state/.style={circle, draw, minimum size=1cm}]]
\node[state] (s1) {$s_1$};
\node[state, right=1.5cm of s1] (s2) {$s_5$};
\node[state, right=1.5cm of s2] (s3) {$s_9$};
\node[state, fill=blue!20, right=1.5cm of s3] (s4) {$s_{13}$};

\node[state, above=1.5cm of s1] (s5) {$s_2$};
\node[state, right=1.5cm of s5] (s6) {$s_6$};
\node[state, right=1.5cm of s6] (s7) {$s_{10}$};
\node[state, fill=red!20, right=1.5cm of s7] (s8) {$s_{14}$};

\node[state, above=1.5cm of s5] (s9) {$s_3$};
\node[state, right=1.5cm of s9] (s10) {$s_7$};
\node[state, right=1.5cm of s10] (s11) {$s_{11}$};
\node[state, right=1.5cm of s11] (s12) {$s_{15}$};

\node[state, above=1.5cm of s9] (s13) {$s_4$};
\node[state, right=1.5cm of s13] (s14) {$s_8$};
\node[state, right=1.5cm of s14] (s15) {$s_{12}$};
\node[state, right=1.5cm of s15] (s16) {$s_{16}$};

\draw[->, line width = 2pt] (s1.south east) -- (s2.south west);
\draw[->] (s2.north west) -- (s1.north east);
\draw[->, line width = 2pt] (s2.south east) -- (s3.south west);
\draw[->] (s3.north west) -- (s2.north east);
\draw[->, line width = 2pt] (s3.south east) -- (s4.south west);

\draw[->, line width = 2pt] (s5.south east) -- (s6.south west);
\draw[->] (s6.north west) -- (s5.north east);
\draw[->, line width = 2pt] (s6.south east) -- (s7.south west);
\draw[->] (s7.north west) -- (s6.north east);
\draw[->, line width = 2pt] (s7.south east) -- (s8.south west);

\draw[->, line width = 2pt] (s9.south east) -- (s10.south west);
\draw[->] (s10.north west) -- (s9.north east);
\draw[->, line width = 2pt] (s10.south east) -- (s11.south west);
\draw[->] (s11.north west) -- (s10.north east);
\draw[->, line width = 2pt] (s11.south east) -- (s12.south west);
\draw[->] (s12.north west) -- (s11.north east);

\draw[->, line width = 2pt] (s13.south east) -- (s14.south west);
\draw[->] (s14.north west) -- (s13.north east);
\draw[->, line width = 2pt] (s14.south east) -- (s15.south west);
\draw[->] (s15.north west) -- (s14.north east);
\draw[->, line width = 2pt] (s15.south east) -- (s16.south west);
\draw[->] (s16.north west) -- (s15.north east);

\draw[->, line width = 2pt] (s5.south west) -- (s1.north west);
\draw[->] (s1.north east) -- (s5.south east);
\draw[->, line width = 2pt] (s9.south west) -- (s5.north west);
\draw[->] (s5.north east) -- (s9.south east);
\draw[->, line width = 2pt] (s13.south west) -- (s9.north west);
\draw[->] (s9.north east) -- (s13.south east);

\draw[->, line width = 2pt] (s6.south west) -- (s2.north west);
\draw[->] (s2.north east) -- (s6.south east);
\draw[->, line width = 2pt] (s10.south west) -- (s6.north west);
\draw[->] (s6.north east) -- (s10.south east);
\draw[->, line width = 2pt] (s14.south west) -- (s10.north west);
\draw[->] (s10.north east) -- (s14.south east);

\draw[->, line width = 2pt] (s7.south west) -- (s3.north west);
\draw[->] (s3.north east) -- (s7.south east);
\draw[->, line width = 2pt] (s11.south west) -- (s7.north west);
\draw[->] (s7.north east) -- (s11.south east);
\draw[->, line width = 2pt] (s15.south west) -- (s11.north west);
\draw[->] (s11.north east) -- (s15.south east);

\draw[->, line width = 2pt] (s12.south west) -- (s8.north west);
\draw[->, line width = 2pt] (s16.south west) -- (s12.north west);
\draw[->] (s12.north east) -- (s16.south east);

\path
    (s4) edge [loop above, line width = 2pt] (s4)
    (s8) edge [loop above, line width = 2pt] (s8);

\end{tikzpicture}
		}
	}
	\caption{Policies computed for the 4$\times$4grid\_avoid benchmark with different randomization modes}
	\label{fig:GridPolicies}
\end{figure}
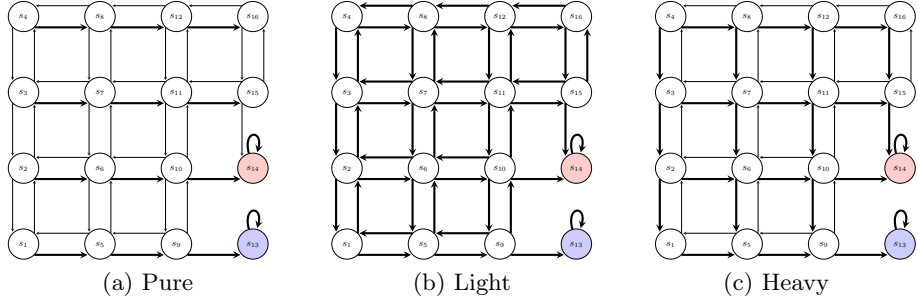

\paragraph{Results.}
Table~\ref{tab:results} summarizes our experimental results. The first column has the name of the benchmark -- for each benchmark, there are four lines representing the four different timeouts used, as specified in the second column (``TO``). Each entry shows both the result (maximum expected discounted reward or maximum reachability probability) for the initial state of the POMDP, and the run time. For each randomization mode (Pure, Light, Heavy) we show both the result of a single MILP call as well as the result of the state splitting heuristic introduced in Sect.~\ref{sec:implementation}, indicated by ``+ H'' in the column name. For the ``+ H'' column, the solve time comprises all calls to the solver as well as the time used for splitting states and observations.\\
Entries printed in \textbf{bold} had the same result and runtime as the optimal solution without timeouts, \ie they were not influenced by the timeouts. In these cases, we omit the entries for higher timeouts, as they had the same values.\\
For entries marked with an star $(\ast)$, the result is either the same as the optimal solution or (for the columns using the splitting heuristic) the same/better than the last iteration that could be solved optimally within two hours.
Additionally, the results for cheese.95 and 4$\times$4.95 are also visualized in Fig.~\ref{fig:Plots} -- the blue dots represent the results for Pure, Light and Heavy randomization modes without state splitting, while the red dots represent the results when applying the splitting heuristic. Data points that have been produced on the same mode, but using different timeouts, are connected by a dashed line.\\
Fig.~\ref{fig:GridPolicies} shows the polices computed by the MILP using different randomization modes for the 4$\times$4grid\_avoid benchmark.

\paragraph{Evaluation.}
Solving the MILP just once and without any randomization is fast, but doesn't yield a very good result in most cases. However, already the ``Light'' randomization can improve the result significantly, in same cases up to a factor of $2$, without significantly increasing the computation time. Adding the full ``Heavy'' randomization yields a further improvement in the result -- most noticeably for the 4$\times$4.95 benchmark, where the result is improved by factor $7$ -- but it can also significantly increase the run time of the solver.\\
While some benchmarks, like 4$\times$4.95 and 4$\times$4grid\_avoid, profit immensely from adding randomization, others, like 1d, have an immediate benefit when using preprocessing. parr95.95 and shuttle.95 even achieve the same results as the reference solvers when applying the state splitting heuristic, while randomization had no effect on the results at all. In general, deterministig, history dependend policies are more powerful than stationary, randomized ones and with arbitrary history, randomization can be simulated -- e.g., taking an action every second time a state is visited.

All benchmarks can achieve results that are very close to those of the reference solvers. In terms of run time, our approach is slower than SARSOP, but highly outperforms solvePOMDP and \prismpomdp.

As can be seen in Fig.~\ref{fig:cheeseplot}, when using different timeouts, the intermediary solution Gurobi returns before fully optimizing the result is already very close to the optimum in many cases. Interestingly, for the mini-hall2 benchmark and the ``Pure + H'' combination, a timeout of 10\,s even yields a better result than a 30\,s timeout: the less-optimized result after 10\,s causes the heuristic to trigger additional state splits that the benchmark ultimately benefits from.

The same cause can result in the values getting worse when more randomization is added, as seen with the 4$\times$4.95 benchmark. The heuristic picks different states to split, resulting in a higher value for the maximum expected discounted reward in the \mbox{Light + H} case than the \mbox{Heavy + H} approach. However, these additional split states also lead to a higher run time.

For the 4$\times$4grid\_avoid benchmark with multi objective model checking, we get UNSAT for the two entries using no randomization, since the required level of reachability probability can not be achieved.

The effects randomization has on a policy are shown in Fig.~\ref{fig:GridPolicies}. All sub-figures depict the 4$\times$4grid\_avoid benchmark, a grid world with one absorbing goal state (blue) and an absorbing bad state (red). All of the white states share an observation and have four possible actions, although we omit the self-loops that occur when trying to move outside of the grid. The arrows corresponsing to actions chosen under the current policy are drawn \textbf{bold}. The system randomly starts in one of the white states.
The policy without randomization always chooses to move right -- only $s_1$, $s_5$ and $s_9$ can reach the goal state. The policy computed with ``Light'' randomization enables all actions for all states -- now each state has the possibility to reach the goal state, but the probability to get to the bad state is still higher.
Only with ``Heavy'' randomization can the bad state be avoided with a higher probability -- each state has equal probability to move down and right, getting the best chance to reach the goal in the lower right corner.

\section{Conclusion}
\label{sec:conclusion}
We introduced a MILP formulation to optimize both reachability probabilities and expected discounted rewards in POMDPs.
We used these MILPs to compute optimal stationary deterministic policies and employed the concept of static randomization.
Furthermore, we introduced state and observation splitting as preprocessing for a POMDP to locally add history to the otherwise stationary policies.
Since blindly splitting states leads to a significant growth of the state space, we proposed a heuristic
that iteratively improves solutions by splitting carefully selected states.
We show the approaches are competitive to state-of-the-art POMDP solvers, and that MILP formulations for rewards and reachability can easily be combined to find policies that satisfy an arbitrary number of specifications at the same time.

\bibliographystyle{splncs04}
\bibliography{abbrev_short,literature}
\end{document}